\documentclass[letterpaper, 10 pt, journal, twoside]{IEEEtran}
\usepackage{cite}
\usepackage{color}
\usepackage{amsmath}
\usepackage{pgfplots,tikz}
\usepackage{graphicx}
\usepackage{caption}
\usepackage{subcaption}
\usepackage{booktabs,multirow,multicol,threeparttable}
\pgfplotsset{compat=1.15}
\newtheorem{lemma}{Lemma}

\begin{document}

\title{Cooperative Coverage with a Leader and a Wingmate in Communication-Constrained Environments}
\author{Sai Krishna Kanth Hari, Sivakumar Rathinam, Swaroop Darbha, David W. Casbeer
\thanks{Sai Krishna Kanth Hari is with the Applied Mathematics and Plasma Physics Division, Los Alamos National Laboratory, NM, 87544 (email: hskkanth@gmail.com).}
\thanks{Sivakumar Rathinam and Swaroop Darbha are with the Department of Mechanical Engineering, Texas A\&M University, College Station, TX 77843.}
\thanks{David W. Casbeer is with the  Autonomous Control Branch, Air Force Research Laboratory, Wright-Patterson A.F.B., OH 45433.}
\thanks{Distribution Statement A: Approved for Public Release; Distribution is Unlimited. PA\# AFRL-2022-4269 and LA-UR-22-30345}
}

\maketitle

\begin{abstract}
We consider a mission framework in which two unmanned vehicles (UVs), a leader and a wingmate,  are required to provide cooperative coverage of an environment while being within a short communication range. This framework finds applications in underwater and/or military domains, where certain constraints are imposed on communication by either the application or the environment. An important objective of missions within this framework is to minimize the total travel and communication costs of the leader-wingmate duo. In this paper, we propose and formulate the problem of finding routes for the UVs that minimize the sum of their travel and communication costs as a network optimization problem of the form of a binary program (BP). The BP is computationally expensive, with the time required to compute optimal solutions increasing rapidly with the problem size. To address this challenge, here, we propose two algorithms, an approximation algorithm and a heuristic algorithm, to solve large-scale instances of the problem swiftly. We demonstrate the effectiveness and the scalability of these algorithms through an analysis of extensive numerical simulations performed over 500 instances, with the number of targets in the instances ranging from 6 to 100.
\end{abstract}

\section{Introduction}
We consider a mission framework in which two unmanned vehicles (UVs), a leader and a wingmate,  are required to provide cooperative coverage of an environment while being within a short communication range. This framework finds applications in underwater and/or military domains, where certain constraints are imposed on communication by either the application or the environment. For example, consider a military application in which UAVs must monitor an environment without being detected. Then, communication using directional antennas is preferred over transmitting omni-directional signals, to reduce the chances of being detected. Alternatively, consider underwater applications such as ocean exploration or mine sweeping. Here, standard terrestrial modes of communication are ineffective due to high attenuation rate of electromagnetic signals, and acoustic mode of communication, which is traditionally used underwater, is undesirable due to a low data transfer date and high latency owing to slow speed of sound in water. As an alternative, the usage of optical signals with a low field of view is preferred for communication, as it offers high data rates and low latency.

In these applications, an unmanned vehicle is accompanied by at least one other vehicle, to either protect and help each other or simply add redundancy to the mission. The vehicles are required to regularly communicate with each other to plan for any unexpected challenges presented by the environment. However, while communicating, the vehicles are required to stay close to each other for multiple reasons. Firstly, staying close to each other reduces the chances of the UVs and their communicating signals being detected in military applications. Secondly, in underwater applications, staying close helps in keeping the optical signals in the low attenuation range. Lastly, staying close reduces the distance required to be traveled by the communicating signals and leads to a reduction of power consumption associated with communication; consequently,  mission-critical tasks such as collecting data through sensors and performing necessary on-board computations will have more on-board power available. Therefore, we consider a problem in which the objective is to route the vehicles such that the combined cost of traveling and communicating is minimized.

In this work, we assume that exactly two UVs are utilized to accomplish the mission. The UVs are required to provide a coverage of the environment by collecting data from representative targets and cover each other by staying close and communicating regularly. Such an assumption is apt for missions in which stealthiness is key and swarming the environment with UVs is undesirable. For convenience of algorithmic development, we assume that the number of targets in the environment is even, say $2m$, and the travel times for the UVs between the targets obey the triangle inequality. Then, we propose the following mission implementation framework. At the beginning of the mission, each UV is assigned a distinct set of $m$ targets to visit and is specified the sequence in which the targets must be visited. Then, the UVs coordinate their speeds such that they reach their respective targets in the specified sequences at the same time. Upon reaching the targets, the UVs quickly exchange information using directional antennas and immediately switch off their communication. Then, they move on to visit the next target in their assigned sequence and repeat the same process of communication until all the targets are exhausted, and they return to their initial location. To realize this framework of coordination and communication, the following problem is of interest:

\emph{Determine the sets of targets assigned to each UV and the sequence (feasible routes) in which the targets in the assigned set must be visited by each UV such that the sum of the travel and communication costs is minimized.} 

Solving this problem is the focus of this work. In this paper, we assume that the cost of communication between UVs present at two targets is nearly equal to the cost incurred by the UVs to travel between the targets, and the travel cost between two targets is proportional to the Euclidean distance between the targets.


\section{Problem Formulation}
We formulate the problem as a network optimization problem of the form of a Binary Program (BP). The targets to be monitored and the travel and communication paths between them are represented by a simple undirected graph ${\mathcal G} = (V,E,c)$; $V$ is the set of vertices representing the targets to be monitored, $E$ is the set of edges representing the travel and communication paths between the targets, $c(.)$, the edge weights, represent the non-negative cost incurred by the UVs to travel and communicate between the targets. Then, a solution to the problem is a connected graph in which all the vertices have a degree $3$; the graph must contain edges that translate to connected paths for each UV to its assigned set of targets and communication links that accord the sequence in which the targets are visited by the UVs.

The mathematical description of the BP that is used to compute such a solution with the minimum travel and communication costs is presented below.

\subsection{Data}
\noindent
$2m$ - Number of targets in the environment\\
$c(i,j)$ - Travel/communication cost between targets $i$ and $j$, where $(i,j) \in E$

\subsection{Variables}
We use sets of binary variables to capture the vehicle's target assignment, and travel and communication paths. First, the set of binary variables $v_i$, $\forall i \in V$, is used to indicate if a target is assigned to the $1^{st}$ UV, i.e., UV-1, or the $2^{nd}$ UV, i.e., UV-2.\\

$v_i$ = $\begin{cases}
1, \text{ if target } i \text{ is assigned to UV-1};\\
0, \text{ otherwise}.
\end{cases}$

Then, we use the set of variables $x_{i,j}$ to indicate whether UV-1 is required to traverse a path $(i,j)$ between targets $i$ and $j$, where $i, j \in V$.\\

$x_{i,j}$ = $\begin{cases}
1, \text{ if edge } (i,j) \in E \text{ is traversed by UV-1};\\
0, \text{ otherwise}.
\end{cases}$

Similarly, we use the set of variables $y_{i,j}$ to indicate whether UV-2 is required to traverse a path $(i,j)$ between targets $i$ and $j$, where $i, j \in V$.\\

$y_{i,j}$ = $\begin{cases}
1, \text{ if edge } (i,j) \in E \text{ is traversed by UV-2};\\
0, \text{ otherwise}.
\end{cases}$

Next, the set of variables $z_{i,j}$ is used to indicate whether an edge $(i,j)$, where $(i,j) \in E$, is utilized by the UVs for communication at any point of time in the mission.\\

$z_{i,j}$ = $\begin{cases}
1, \text{ if edge } (i,j) \text{ is used for communication};\\
0, \text{ otherwise}.
\end{cases}$

Finally, we utilize the sets of variables $\xi_{i,j,l}$ and $\eta_{i,k,l}$, $\forall i,j \in V$, and $l,k \in V \setminus{\{i,j\}}$, to ensure that the communication links are chosen based on the sequence in which the targets are visited by the UVs. \\

$\xi_{i,j,l}$ = $\begin{cases}
1, \text{ if } x_{i,j} = 1 \text{ and } z_{j,l} = 1;\\
0, \text{ otherwise}.
\end{cases}$

$\eta_{i,k,l}$= $\begin{cases}
1, \text{ if } z_{i,k} = 1 \text{ and } y_{k,l} = 1;\\
0, \text{ otherwise}.
\end{cases}$
\subsection{Constraints}
Using these binary variables, the problem requirements can be represented as mathematical constraints as follows.
\subsubsection{Target Assignment} Each UV must be assigned exactly $m$ targets. This is represented by Equation \eqref{eq:m-targets}.
    \begin{equation}
        \sum_{i \in V} v_i = m \label{eq:m-targets}
    \end{equation}

\subsubsection{Feasible Tours/Sequences}
A UV can traverse between two targets (along an edge with the targets as the endpoints) only if both the targets are assigned to the UV. Constraints \eqref{ineq:target-i-picked-edge-x-picked}--\eqref{ineq:target-j-picked-edge-x-picked} ensure that an edge, $(i,j) \in E$, is assigned to UV-1's tour only if targets $i$ and $j$ are both assigned to UV-1, and constraints \eqref{ineq:target-i-picked-edge-y-picked}--\eqref{ineq:target-j-picked-edge-y-picked} ensure that an edge $(i,j)$ is assigned to UV-2's tour only if both $i$ and $j$ are assigned to UV-2.

\begin{align}
    x_{i,j} \leq v_i, \quad \forall (i,j) \in E \label{ineq:target-i-picked-edge-x-picked}\\
    x_{i,j} \leq v_j, \quad \forall (i,j) \in E \label{ineq:target-j-picked-edge-x-picked}\\
    y_{i,j} \leq 1-v_i, \quad \forall (i,j) \in E \label{ineq:target-i-picked-edge-y-picked}\\
    y_{i,j} \leq 1-v_j, \quad \forall (i,j) \in E \label{ineq:target-j-picked-edge-y-picked}
\end{align}

The purpose of constraints \eqref{eq:x-picked-if-target-picked}--\eqref{eq:y-picked-if-target-picked} is similar to that of \eqref{ineq:target-i-picked-edge-x-picked}--\eqref{ineq:target-j-picked-edge-y-picked}, as they ensure that a UV can only travel to targets that are assigned to it. 

\begin{align}
    \sum_{i \in V\setminus\{j\}} x_{i,j} = v_j, \quad \forall j \in V \label{eq:x-picked-if-target-picked}\\
    \sum_{i \in V\setminus\{j\}} y_{i,j} = 1 - v_j, \quad \forall j \in V \label{eq:y-picked-if-target-picked}
\end{align}

Constraints \eqref{eq:uv1-enter-leave} and \eqref{eq:uv2-enter-leave} ensure that a UV arriving at a target must also depart from the target.

\begin{align}
    \sum_{i \in V\setminus\{j\}} x_{i,j} = \sum_{l \in V\setminus\{j\}} x_{j,l}, \quad \forall j \in V \label{eq:uv1-enter-leave}\\
    \sum_{i \in V\setminus\{j\}} y_{i,j} = \sum_{l \in V\setminus\{j\}} y_{j,l}, \quad \forall j \in V \label{eq:uv2-enter-leave}
\end{align}

In addition to the aforementioned degree/flow requirements, a UV's path is required to be connected and must not contain sub-tours. This can be enforced by constraints \eqref{ineq:ste-1}--\eqref{ineq:ste-2}, which are referred to as sub-tour elimination constraints.
\begin{align}
    \sum_{i \in S, j \notin S} x_{i,j} \geq v_s, \quad \forall s \in S, S \subset V, |S| \leq m-1, \label{ineq:ste-1}\\
    \sum_{i \in S, j \notin S} y_{i,j} \geq 1 - v_s, \quad \forall s \in S, S \subset V, |S| \leq m-1. \label{ineq:ste-2}
\end{align}

\subsubsection{Communication Links} Once the UVs travel paths are determined, the communication links can be described using the set of constraints \eqref{ineq:comm-once-at-target} -- \eqref{cons:last}. Firstly, the UVs communicate only once at every target. This is expressed by the set of inequalities \eqref{ineq:comm-once-at-target}.

\begin{align}
    \sum_{j \in V \setminus\{i\}} z_{i,j} \leq 1, \quad \forall i \in V \label{ineq:comm-once-at-target}
\end{align}

Because every UV visits exactly $m$ distinct targets, there are a total of $m$ communication links; this is captured by Equation \eqref{eq:m-comm-links}.
\begin{equation}
    \sum_{(i,j) \in E} z_{i,j} = m \label{eq:m-comm-links}
\end{equation}

Then, because a communication is only performed between two different vehicles, there are no communication links between targets assigned to the same UV. This is captured by the sets of inequalities described by \eqref{ineq:no-comm-bet-uv1} and \eqref{ineq:no-comm-bet-uv2}.
\begin{align}
    z_{i,j} \leq v_i + v_j, \quad \forall (i,j) \in E \label{ineq:no-comm-bet-uv1}\\
    z_{i,j} \leq 2 - (v_i + v_j), \quad \forall (i,j) \in E \label{ineq:no-comm-bet-uv2}
\end{align}

Next, it must be ensured that the assignment of communication links is consistent with the sequence in which the targets are visited by the UVs. This requirement can be restated in the form of the following lemma.\\
    
    \begin{lemma}
    \label{lemma:order-preserving}
     Consider any 2 arbitrary and distinct vertices $i$, $l \neq i$ $\in V$. Then, there exists a vertex $j \in V\setminus\{i,l\}$ such that $(i,j)$ belongs to UV-1's tour and $(j,l)$ is a valid communication link if and only if there exists a vertex $k\in V\setminus\{i,l\}$ such that $(k,l)$ belongs to UV-2's tour and $(i,k)$ is a valid communication link. In other words, $j \in V\setminus\{i,l\}$ such that $\xi_{i,j,l} = 1$ iff $k\in V\setminus\{i,l\}$ such that $\eta_{i,k,l} = 1$.
    \end{lemma}
    
    Lemma \ref{lemma:order-preserving} can be enforced using the following set of constraints.
    
    \begin{align}
        \sum_{j \in V\setminus\{i,l\}}\xi_{i,j,l} = \sum_{k \in V\setminus\{i,l\}}\eta_{i,k,l}, \quad ,\forall i, l \in V\\
        \xi_{i,j,l} \leq x_{i,j}, \quad \forall i \in V, j \in V \setminus\{i\}, l \in V \setminus\{j\} \\
        \xi_{i,j,l} \leq z_{j,l}, \quad \forall i \in V, j \in V \setminus\{i\}, l \in V \setminus\{j\} \\
        \xi_{i,j,l} \geq x_{i,j} + z_{j,l} - 1, \quad \forall i \in V, j \in V \setminus\{i\}, l \in V \setminus\{j\} \\
        \eta_{i,k,l} \leq z_{i,k}, \quad \forall i \in V, k \in V \setminus\{i\}, l \in V \setminus\{k\} \\
        \eta_{i,k,l} \leq y_{k,l}, \quad \forall i \in V, k \in V \setminus\{i\}, l \in V \setminus\{k\} \\
        \eta_{i,k,l} \geq z_{i,k} + y_{k,l} - 1, \quad \forall i \in V, k \in V \setminus\{i\}, l \in V \setminus\{k\} \label{cons:last}
    \end{align}

\subsection{Objective}
The objective of the problem is to find feasible tours and communication links that minimize the
total travel and communication costs incurred by the UVs. This translates to minimizing the sum of the costs of all the selected edges and can be mathematically represented by \eqref{exp:obj}.
\begin{align}
    \min \sum_{(i,j) \in E} c_{i,j}  (x_{i,j} + y_{i,j} + z_{i,j}) \label{exp:obj}
\end{align}

The objective function \eqref{exp:obj}, together with constraints \eqref{eq:m-targets}--\eqref{cons:last}, defines the Binary Program of interest.

\section{Computing Optimal Solutions}
\subsection{Implementation}
We implemented the binary program in Julia, using JuMP, a package for mathematical optimization. For every given instance, we first evaluate the function $c(i,j)$, $\forall (i,j) \in E$, by calculating the Euclidean distances between the targets. Then, we compute optimal solutions to the problem by solving the binary program using Gurobi \cite{gurobi}, a commercial off-the-shelf optimization solver that is known to be one of the best in handling binary programs.

It is to be noted that the number of constraints specified by the inequalities \eqref{ineq:ste-1} and \eqref{ineq:ste-2} is an exponential function of $m$, and directly adding such a large set of constraints to the problem makes solving it difficult. Therefore, we adopt a lazy callback approach \cite{lazy}, where we first solve the problem without the sub-tour elimination constraints, and then iteratively add only the violated sub-tour elimination constraints and re-solve the problem until a solution that satisfies all the constraints is obtained; this solution is optimal to the problem.


\subsection{Computation Time} \label{subsec:comp-time}

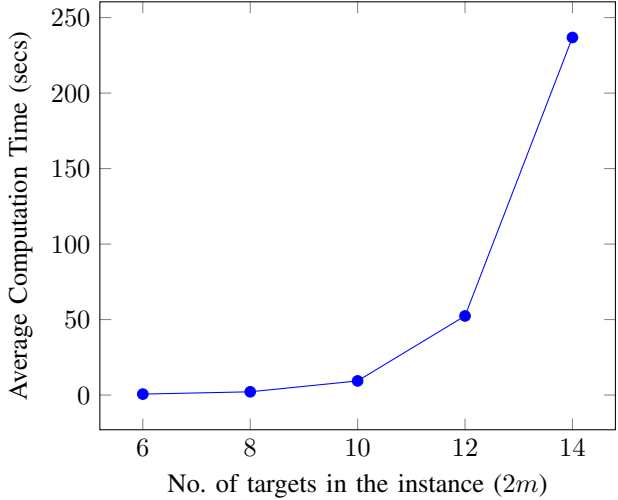
\begin{figure}[h!]
    \centering
   \begin{tikzpicture}
        \begin{axis}[
            xlabel= No. of targets in the instance ($2m$),
        	ylabel= Average Computation Time (secs),
        	legend pos = north west,
        	scale=1
        ]
        \addplot[
            color=blue,
            solid,
            mark=*,
            mark options={solid}
            ]
            coordinates {
            (6,0.64)(8,2.16)(10,9.37)(12,52.41)(14,236.84)
         };
        \end{axis}
    \end{tikzpicture}
    \caption{Figures depicts the rapid increase in the average computation time required to solve the problem with the number of targets in the instance; the average is taken over $50$ instances each, for a given number of targets.}
    \label{fig:avg-opt-comp-time}
\end{figure}

We used this implementation to solve the problem for 300 randomly generated instances, with the number of targets in these instances ranging from $6$ to $16$; there are 50 instances each with $2m = 6$, $8$, $10$, $12$, $14$, and $16$. The simulations suggest that the average computation time required to solve the problem increases rapidly with the number of targets in the instance; the average computation time (taken over $50$ instances) against the number of targets in the instance is shown in Figure \ref{fig:avg-opt-comp-time}. The solver was  unsuccessful in providing optimal solutions for instances with $2m \geq 16$ within a 3 hour cut-off time.

Therefore, solving the BP to compute optimal solutions for large-scale instances is impractical. Thus, in this work, we develop approximate and heuristic algorithms to compute good quality solutions to the problem swiftly.

\section{Approximation Algorithm}
An approximation algorithm provides solutions with a worst-case guarantee on its quality. The quality of the solution is determined by approximation ratio, which is the ratio of the worst-case cost of the solution provided by the algorithm to the cost of the optimal solution to the problem. Furthermore, the computation time required for implementing an approximation algorithm must be a polynomial function of the input size i.e., $m$. Here, we develop such an algorithm with an approximation ratio of $3.75$.

\subsection{Algorithm}
\begin{figure}
     \centering
     \begin{subfigure}[b]{0.3\textwidth}
         \centering
         \includegraphics[width=\textwidth]{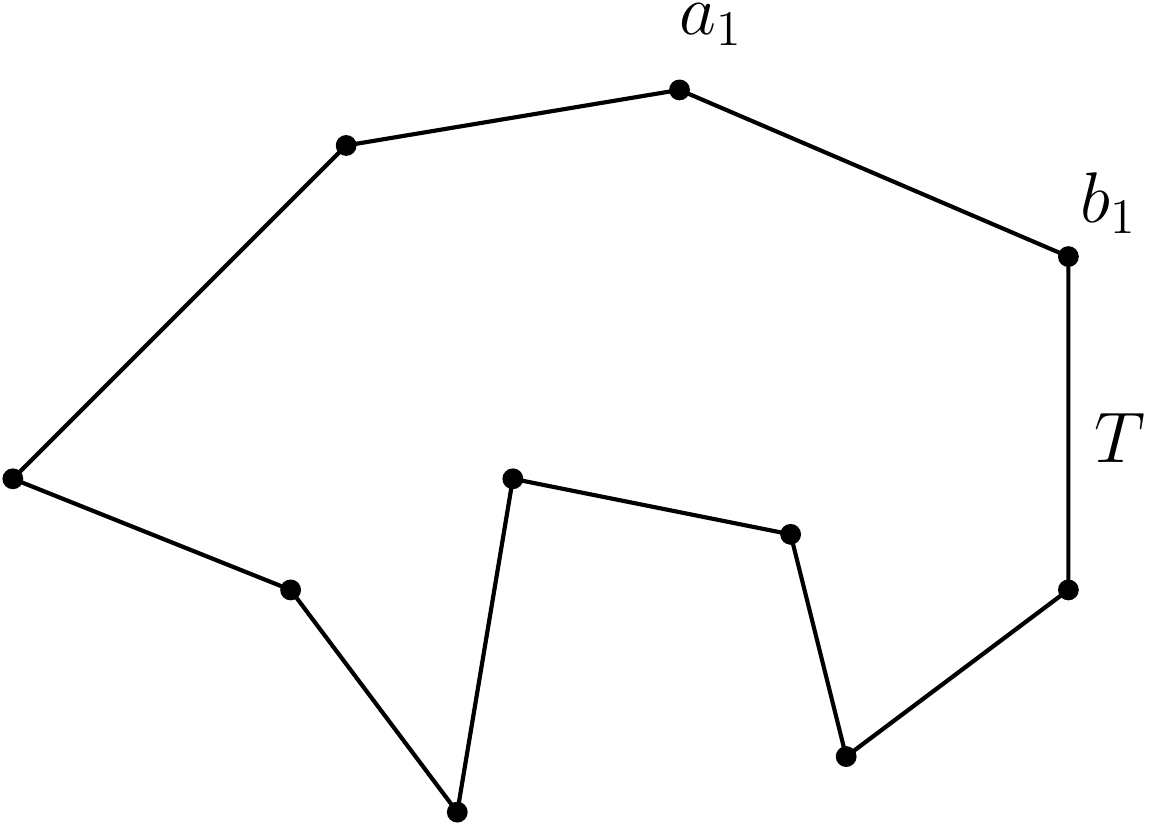}
         \caption{Single UV tour, $T$, shown in black}
         \label{fig:aa-chris-tour}
     \end{subfigure}
     \hfill
     \begin{subfigure}[b]{0.3\textwidth}
         \centering
         \includegraphics[width=\textwidth]{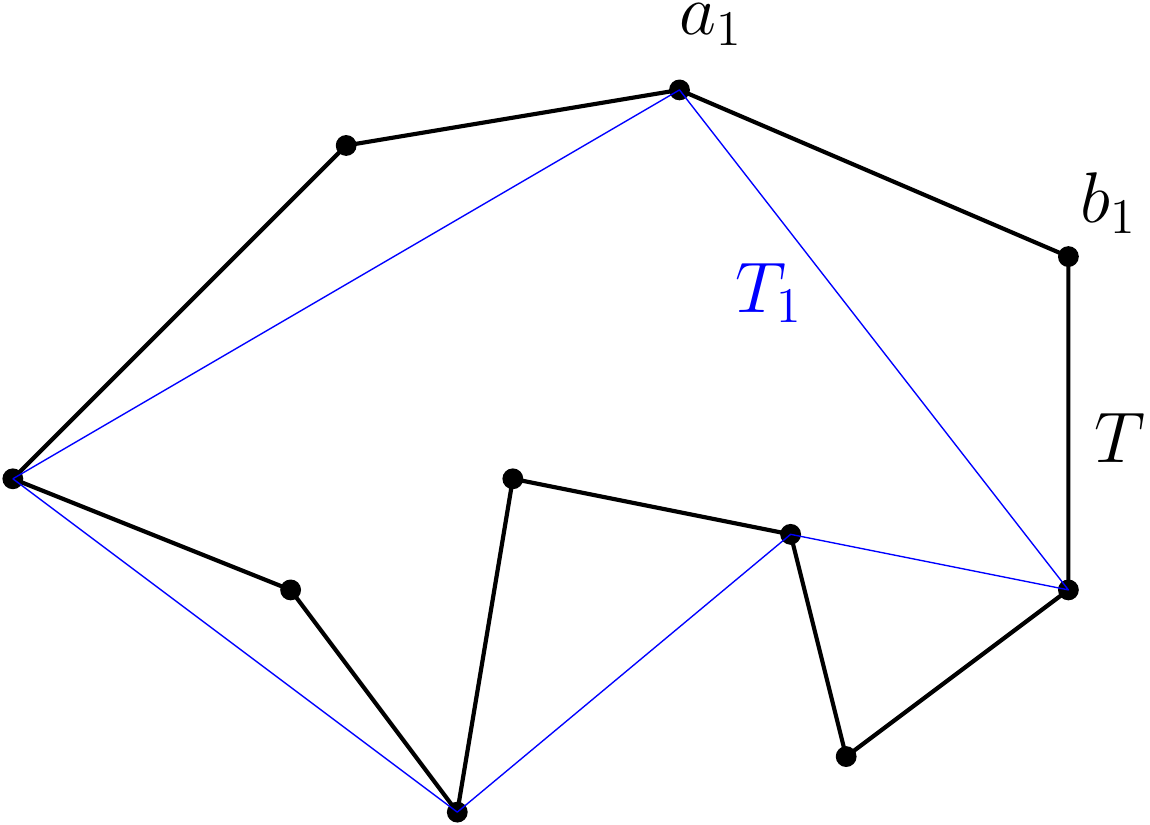}
         \caption{$1^{st}$ UV's tour, $T_1$, shown in blue}
         \label{fig:aa-v1-tour}
     \end{subfigure}
     \hfill
     \begin{subfigure}[b]{0.3\textwidth}
         \centering
         \includegraphics[width=\textwidth]{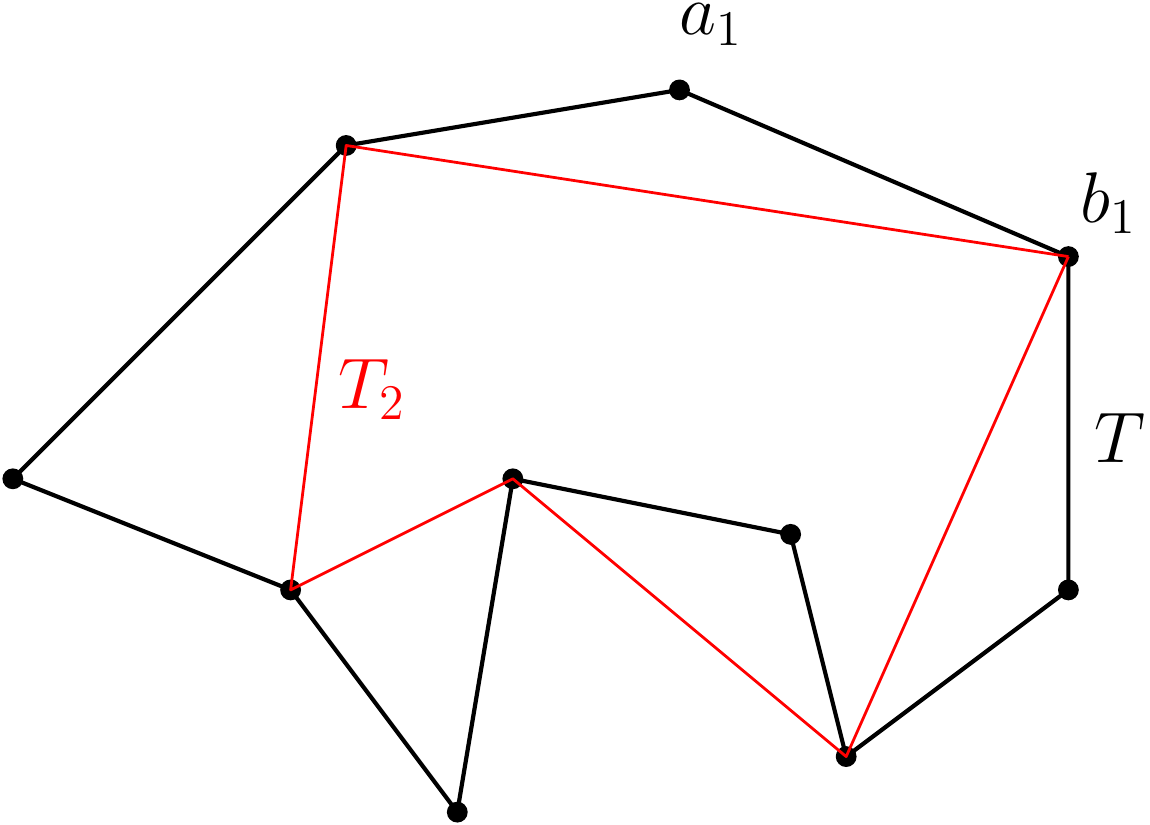}
         \caption{$1^{st}$ UV's tour, $T_2$, shown in red}
         \label{fig:aa-v2-tour}
     \end{subfigure}
     \hfill
     \begin{subfigure}[b]{0.3\textwidth}
         \centering
         \includegraphics[width=\textwidth]{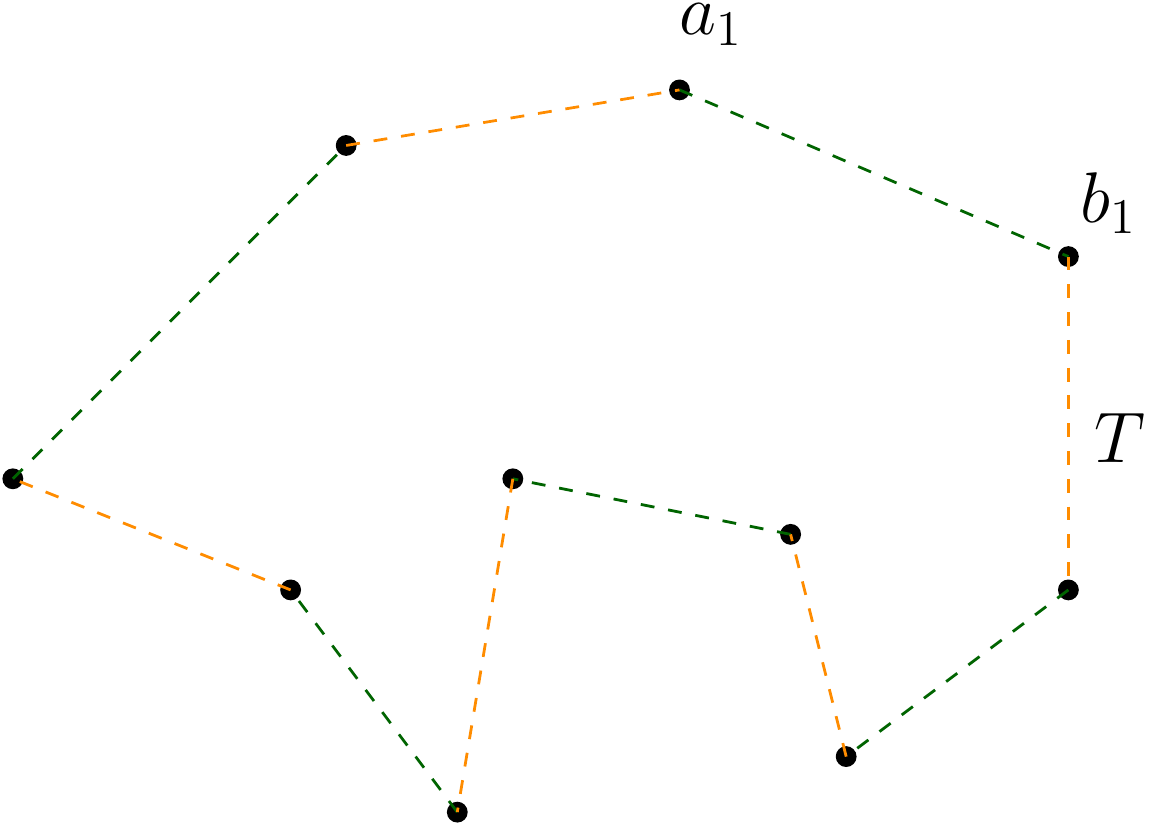}
         \caption{Potential sets of communication links shown in orange and green}
         \label{fig:aa-pre-comm}
     \end{subfigure}
    \hfill
     \begin{subfigure}[b]{0.3\textwidth}
         \centering
         \includegraphics[width=\textwidth]{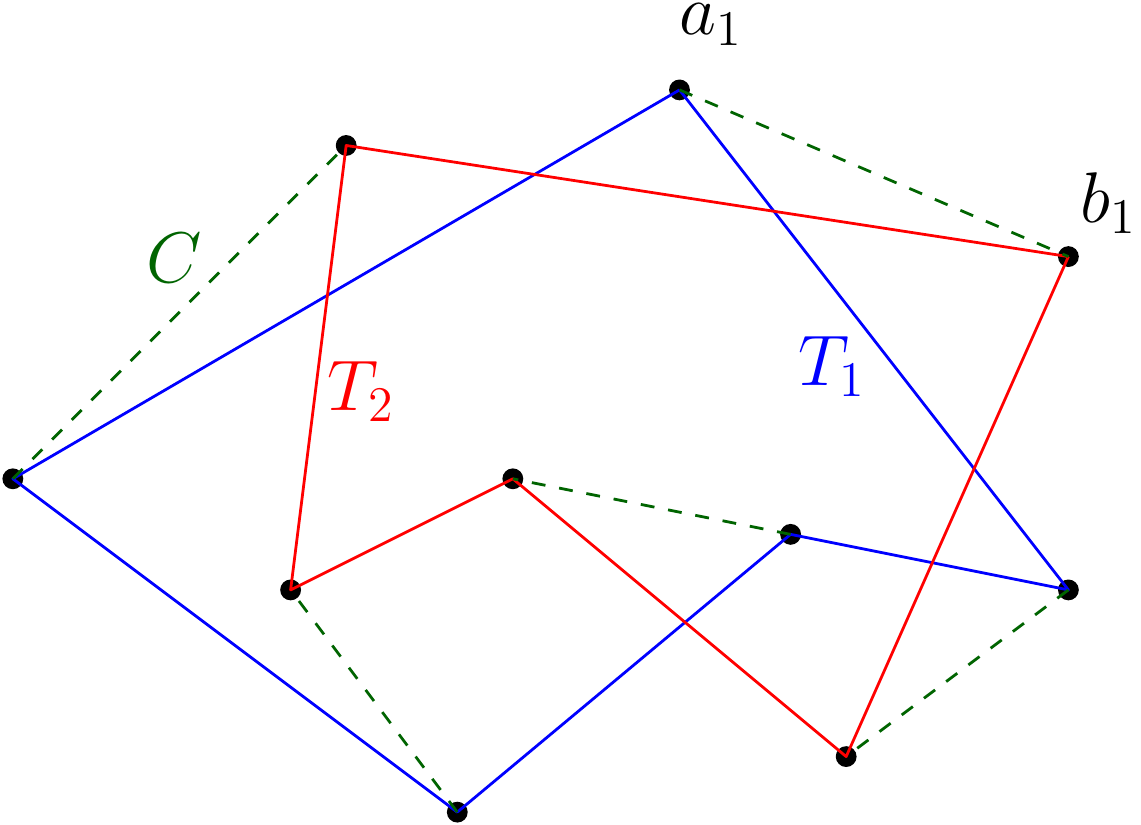}
         \caption{Final solution of the approximation algorithm, containing feasible tours $T_1$ (blue) and $T_2$ (red), and the communication links, $C$ (green).}
         \label{fig:aa-solution}
     \end{subfigure}
     \caption{Figure illustrates the steps of the approximation algorithm on a $10$-target instance.}
     \label{fig:aa-figure}
\end{figure}

We illustrate the steps of the algorithm on a $10-$target instance shown in Figure \ref{fig:aa-figure}. The first step of the algorithm involves computing a single UV tour over all the targets in the environment, i.e., a tour that allows a single UV to start at a target, visit all the targets once, and return to the starting point. Such a tour of the minimum travel cost is an optimal solution to the classic Traveling Salesman Problem (TSP). Here, we utilize the well-known Christofides algorithm \cite{christofides1976worst}, an approximation algorithm to the TSP with an approximation ratio of 1.5, to compute the desired single UV tour. Let this tour be denoted by $T$; figure \ref{fig:aa-chris-tour} illustrates a single UV tour spanning all the $10$ targets. We tailor this tour to construct the desired feasible tours for both the UVs and the communication links between them.


The next step of the algorithm provides feasible tours for both the UVs in the mission. To obtain the tour for the $1^{st}$ UV, pick any vertex, say $a_1$, from $T$. Then, starting at $a_1$, construct a graph by joining only the alternate vertices of $T$, as shown by the blue edges in Figure \ref{fig:aa-v1-tour}; denote the constructed graph by $T_1$. In other words, $T_1$ is a tour obtained by skipping visits to alternate vertices from $T$. $T_1$ is the desired tour for the $1^{st}$ UV in the mission, covering half of the targets in the environment. Now, consider the vertex $b_1$, adjacent to $a_1$, in $T$. Then, similar to the construction of $T_1$, construct a graph starting at $b_1$ and skipping alternate vertices of $T$, as shown in Figure \ref{fig:aa-v2-tour}; denote the obtained graph by $T_2$. $T_2$ represents the desired tour for the $2^{nd}$ UV in the mission, covering the rest of the targets.



The final step of the algorithm involves identifying the communication links between the vehicles when they visit their respective targets. Thusfar, we have feasible tours for both the UVs, but the starting points of the UVs in their tours is not yet determined. This leaves us with a number of options to choose the communication links from. In this algorithm, we restrict the communication links to the set of edges of $T$. In that case, the two sets of alternating edges of $T$ are candidate communication links; either the set containing the edge $a_1-b_1$, shown in green in Figure \ref{fig:aa-pre-comm}, or the set that does not contain this edge, shown in orange in Figure \ref{fig:aa-pre-comm}. From these two sets, we choose the set that has the minimum sum of the costs of all its edges. Let this set of edges be $C$. Then, $T_1$, $T_2$, and $C$ together constitute the solution of the algorithm and it is a feasible solution to the problem; solution of the algorithm for the illustrative instance is shown in Figure \ref{fig:aa-solution}.

It is to be noted that all the steps of the algorithm can be performed in a polynomial time. Next, to obtain worst-case guarantees, we need to analyze the cost of the solution obtained from this algorithm.

\subsubsection{Cost of solutions} 
The cost of the solution of this algorithm is the sum of the costs of the edges of $T_1$, $T_2$, and $C$, i.e., $Cost(Solution) = Cost(T_1) + Cost(T_2) + Cost(C)$.
Let us analyze the edge costs of these graphs individually. Firstly, recall that the sum of the cost of edges of $T$ is at most $\frac{3}{2} TSP^*$, where $TSP^*$ is the cost of an optimal TSP tour over the targets. This follows from the approximation ratio of the Christofide's tour $T$ \cite{christofides1976worst}. Then, recall that $T_1$ and $T_2$ are obtained by skipping or shortcutting visits from $T$. Then, due to the triangle inequality of edge costs, it follows that the sum of the costs of edges of $T_1$ is at most that of $T$. That is, $Cost(T_1) \leq Cost(T)$. Since the construction of $T_2$ is similar to that of $T$, we have that $Cost(T_2) \leq Cost(T)$.

Now, recall that $C$ is the set of alternating edges of $T$ that has the minimum cost. Let the costs of alternating edges of $T$ aggregate to $c_1$ and $c_2$. Then, $c_1 + c_2 = Cost(T)$. WLOG, suppose that $c_1 \leq c_2$. Then, we have that $Cost(C) = c_1$. Combining these, it follows that 
\begin{align*}
    Cost(T) &= c_1 + c_2\\
            &\geq c_1 + c_1\\
            &= 2*Cost(C)
\end{align*}

Therefore, $Cost(C) \leq \frac{1}{2}Cost T$. Consequently, the cost of the solution provided by the algorithm can be upper bounded as follows
\begin{flalign*}
    Cost(Solution) &= Cost(T_1) + Cost(T_2) + Cost(C)\\
    &\leq \frac{5}{2}Cost(T)\\
    &\leq \frac{5}{2} *\frac{3}{2} TSP^*\\
    &= \frac{15}{4} TSP^*
\end{flalign*}

\subsection{Lower Bound}
\begin{figure}
     \centering
     \begin{subfigure}[b]{0.3\textwidth}
         \centering
         \includegraphics[width=\textwidth]{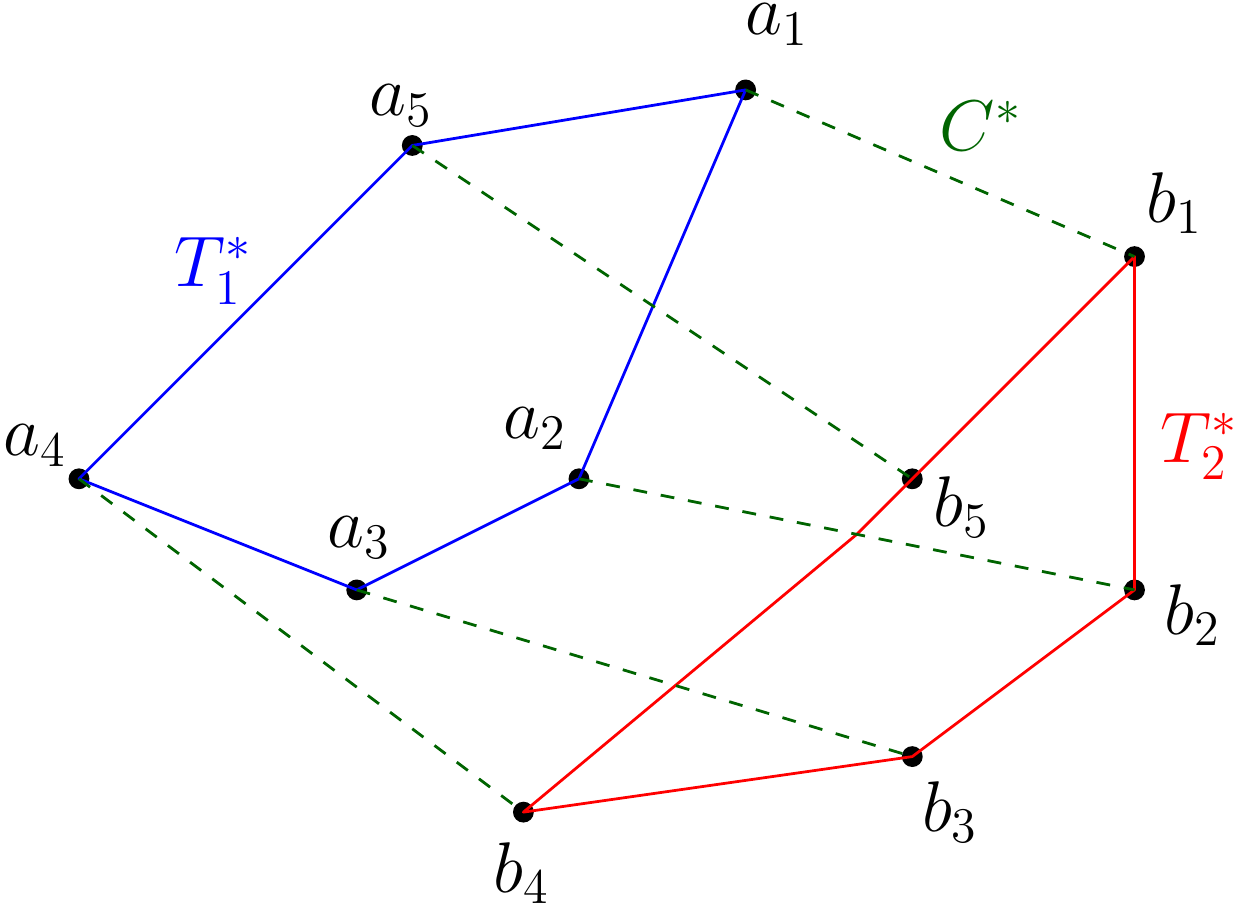}
         \caption{Optimal solution, $O^*$, used to illustrate the development of the lower bound}
         \label{fig:aa-lb-1}
     \end{subfigure}
     \hfill
     \begin{subfigure}[b]{0.3\textwidth}
         \centering
         \includegraphics[width=\textwidth]{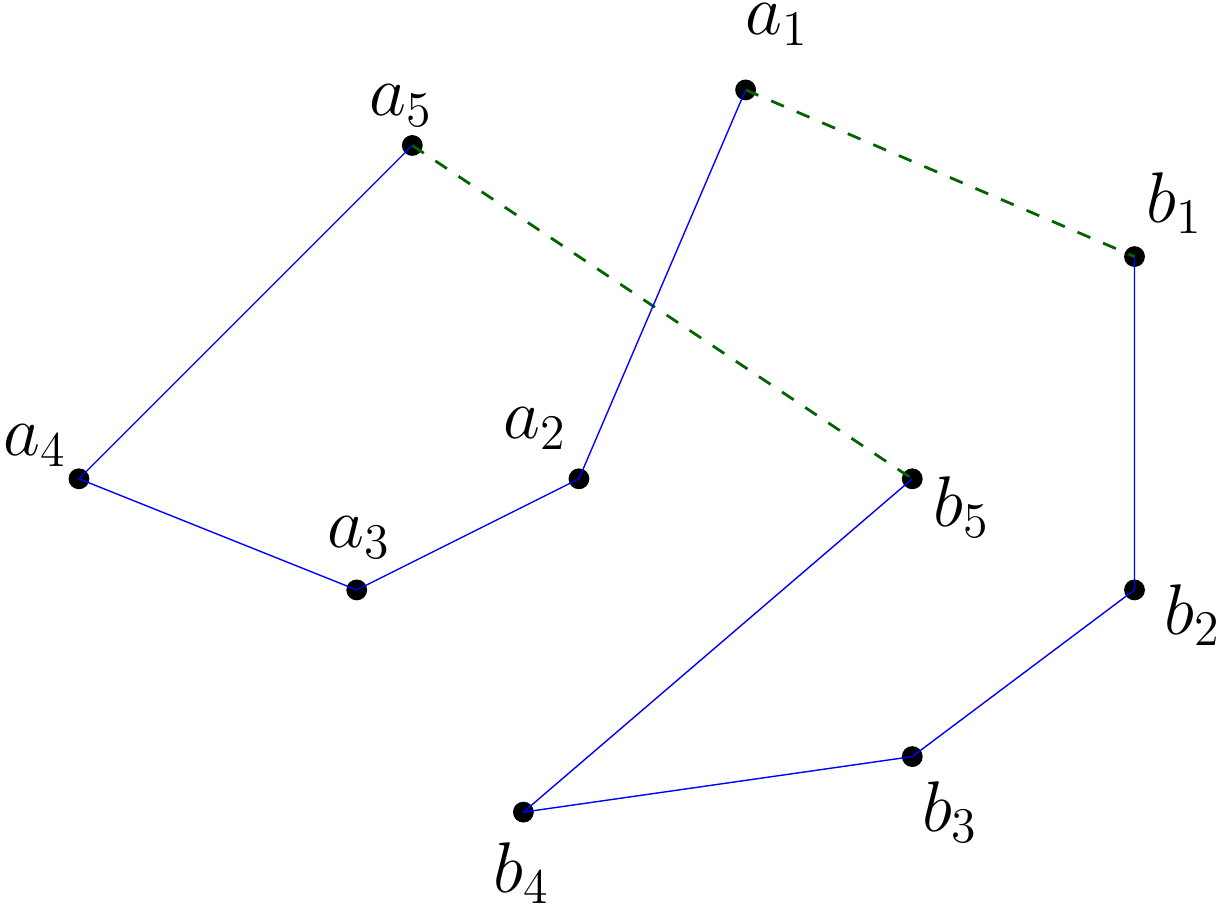}
         \caption{Single UV tour over all the targets obtained by removed a few edges from $O^*$}
         \label{fig:aa-lb-2}
     \end{subfigure}
     \hfill
     \begin{subfigure}[b]{0.3\textwidth}
         \centering
         \includegraphics[width=\textwidth]{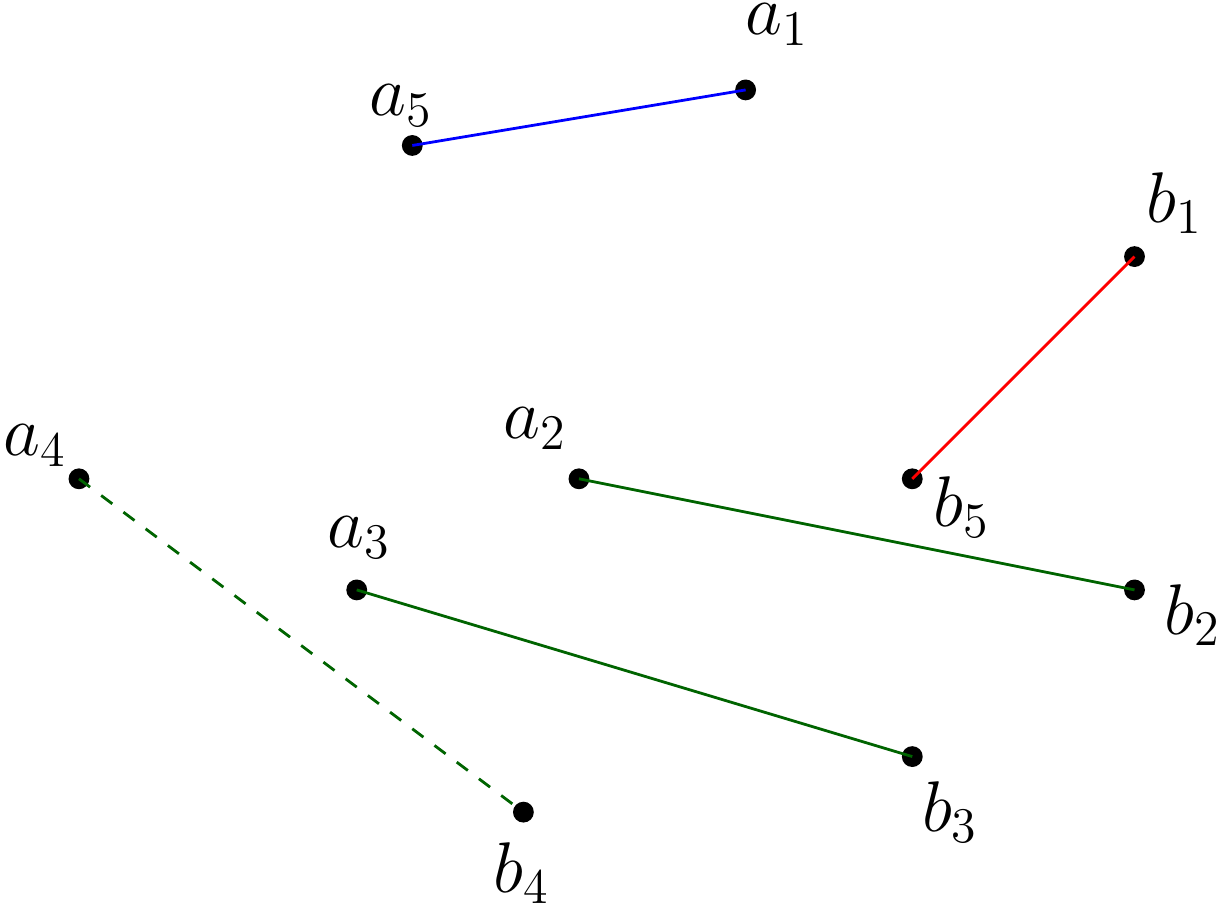}
         \caption{Perfect bipartite matching formed by the remaining edges of $O^*$}
         \label{fig:aa-lb-3}
     \end{subfigure}
     \caption{Figures illustrating the development of the lower bound to the problem.}
     \label{fig:lb-illustration}
\end{figure}

To arrive at an approximation ratio for the algorithm, we develop a lower bound to the optimal solution of the problem. Towards this end, let $O^* = T_1^* \cup T_2^* \cup C^*$ be an optimal solution, where $T_1^*$ and $T_2^*$ are the tours obtained by traversing the targets in the orders specified by the following tuples respectively: $(a_1, a_2, \dots, a_m, a_1)$, $(b_1, b_2, \dots, b_m, b_1)$, and $C^* = \{(a_1,b_1), \dots, (a_m, b_m) \}$ denotes the set of communication links, as shown in Figure \ref{fig:aa-lb-1}. Now, from $O^*$, consider only the edges obtained by traversing across all the vertices in the specified order: $(a_1, a_2, \dots, a_m, b_m, b_{m-1}, \dots, b_1,a_1)$.
These edges form a single UV tour over all the targets, as shown in Figure \ref{fig:aa-lb-2}. Therefore, the cost of these edges is at least $TSP^*$. Next, observe that the rest of the edges in $O^*$, i.e., $\{(a_2,b_2), (a_3, b_3), \dots, (a_{m-1},b_{m-1}),(a_1,a_m), (b_1,b_m)\}$, form a perfect bi-partite matching, as shown in Figure \ref{fig:aa-lb-3}. Therefore, the cost of these edges is at least equal to $Matching^*$, which is the cost of the minimum weighted perfect bi-partite matching of $G$. As a result, $TSP^* + Matching^*$ is a valid lower bound for the cost of the optimal solution.

From above, we have that $Cost(Solution) \leq \frac{15}{4} TSP^*$, and here, we have that $Cost(Optimal \text{ } Solution) \geq TSP^* + Matching^*  \geq TSP^*$. Therefore, $3.75$ $(=\frac{15}{4})$ is an approximation ratio of the algorithm. This ratio is also referred to as the apriori ratio.

\subsection{Heuristics}
If the solve time being a polynomial function of the input size is not a requirement, in the first step of the algorithm, one can use an alternative single UV tour with a lower cost. For example, one can use a tour $T$ obtained from the Lin-Kernighan heuristic \cite{lin1973effective}, which is known to provide near-optimal solutions to the TSP for many instances quickly. This helps in improving the quality of the solutions.

\section{Results}
In this section, we evaluate the performance of the proposed Heuristic and Approximation algorithms by presenting the results of numerical simulations performed on 500 instances of the problem. The instances are created by randomly generating target coordinates over a 500 $\times$ 500 grid (using the rand function in Julia \cite{bezanson2017julia}). The number of targets in these instances range from 6 to 100, with exactly 50 instances each with $2m = 6$, $8$, $10$, $12$, $14$, $20$, $30$, $40$, $50$, and $100$. 
We evaluate the performance of these algorithms using two criteria: 1) the quality of the solutions provided by the algorithm, 2) the time taken to compute these solutions. A desirable algorithm provides good-quality solutions quickly.

All the computations presented in this section were performed on a MacBook Pro with 16 GB RAM, Quad-Core Intel Core i7 processor @ 2.8 GHz. All the implementations, except for the computation of the LKH tours, were performed in the language Julia. For the computation of optimal solutions, a Julia package for mathematical programming, JuMP \cite{DunningHuchetteLubin2017}, and a commercially available optimization solver Gurobi \cite{gurobi}, were utilized. LKH tours were computed using the software package \cite{LKH-solver}.

To evaluate the quality of the solutions, we use metrics that help us in understanding the gap between the cost of the algorithmic solution and the optimal cost for each instance. Towards this end, we utilize the aposteriori ratio of algorithmic solutions, which is the ratio of the cost of the algorithmic solution to the optimal cost; note that this can be different from the approximation ratio (apriori ratio), which is a worst-case theoretical bound on the aposteriori ratio). Nonetheless, as discussed in Section \ref{subsec:comp-time}, it is difficult to compute optimal solutions to large instances. Therefore, we also use another metric to aid our understanding of the quality of the solution. This metric is the ratio of the cost of the algmorithmic solution to the lower bound on the optimal cost. While this metric does not directly indicate the gap of the cost of the algorithmic solution from the optimal cost, it provides a bound on the same for instances where the optimal cost is not available. 

For the reasons discussed above, we categorize the set of all instances into two categories:  1) small instances, which contain the instances for which optimal solutions are available, and 2) large instances, which contain the instances for which optimal solutions were not computable within a 3-hour threshold. The former category contains 250 instances, with the number of targets ranging from 6 to 14, whereas the latter contains the remaining 250 instances, with the number of targets ranging from 20 to 100.

\subsection{Small Instances}
For every instance in this category, we compute the optimal solution, heuristic solution, the approximate solution, and a lower bound to the optimal value using the procedures discussed above. Then, for the heuristic and approximate solutions, we compute the aposteriori ratios and the ratios of their costs to the lower bound. In Table \ref{tab:avg-ratios-small-instances}, we present the averages of each of these ratios, where the average is computed over ($50$) instances with identical number of targets. From the table, it can firstly be seen that the approximation algorithm provides much better solutions in practice compared to the proposed worst-case bounds (apriori ratios). While the apriori ratio of the algorithm is 3.75, the aposteriori ratios were observed to range only from 1.05 to 1.12. Secondly, barring the exception of the $6-$target instances, the aposteriori ratios of the approximation algorithm were observed to increase with the number of targets. While a similar pattern was observed in the case of the heuristic algorithm, the rate of increase was lower in this case. Thirdly, once again barring the exception of the $6-$target instances, the aposteriori ratios of the heuristic algorithm were smaller than those of its counterpart. Therefore, if the presence of a polynomial time algorithm and a worst-case bound is not a requirement, the heuristic algorithm tends to provide better quality solutions for small instances.

In addition to the aposteriori ratios, the table also includes the ratios of the costs of algorithmic solutions to the lower bounds of the optimal values. These ratios help in evaluating the quality of the algorithmic solutions in the absence of optimal solutions. While optimal solutions are available in this case, we utilize these values to benchmark and understand the quality of the solutions for large instances. It can be observed from the table that these values are higher than their corresponding aposteriori ratios. The ratios to the lower bounds range from 1.39 to 1.55 for the approximation algorithm and from 1.40 to 1.44 for the heuristic algorithm. These relatively higher ratios can be attributed to the slack in the lower bounds. 

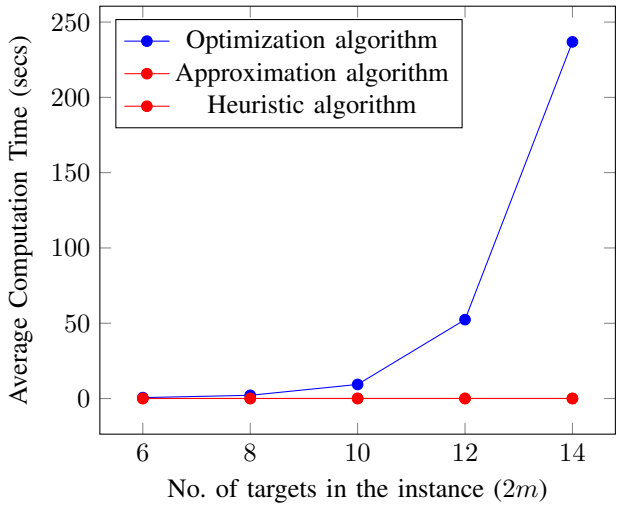
\begin{figure}[htb]
    \centering
   \begin{tikzpicture}
        \begin{axis}[
            xlabel= No. of targets in the instance ($2m$),
        	ylabel= Average Computation Time (secs),
        	legend pos = north west,
        	scale=1
        ]
        \addplot[
            color=blue,
            solid,
            mark=*,
            mark options={solid}
            ]
            coordinates {
            (6,0.64)(8,2.16)(10,9.37)(12,52.41)(14,236.84)
         };
         \addlegendentry{Optimization algorithm}
                 \addplot[
            color=red,
            solid,
            mark=*,
            mark options={solid}
            ]
            coordinates {
            (6,0.06)(8,0.06)(10,0.06)(12,0.06)(14,0.06)
         };
            \addlegendentry{Approximation algorithm} 
            \addplot[
            color=red,
            solid,
            mark=*,
            mark options={solid}
            ]
            coordinates {
            (6,0.05)(8,0.05)(10,0.05)(12,0.05)(14,0.05)
         };
         \addlegendentry{Heuristic algorithm}
        \end{axis}
    \end{tikzpicture}
    \caption{The figure depicts the average computation time required to solve the optimization, approximation, and heuristic algorithms in blue, red and red colors respectively. The averages are computed over $50$ instances each for a given number of targets.}
    \label{fig:avg-algo-comp-time}
\end{figure}

The next criteria in evaluating the algorithms is their computation time. In Figure \ref{fig:avg-algo-comp-time}, we plot the average time required to compute the approximate, heuristic and the optimal solutions against the number of targets in the instance. The figure indicates that the computation times for the approximate and heuristic algorithms are similar. However, these times are significantly lower than that required to compute optimal solutions. Therefore, these algorithms help in determining quick solutions to the problem for large instances. In Table \ref{tab:sample-values-small-instances}, we present the results of the simulations for $50$ illustrative small instances.

\begin{table*}[]
    \caption{The table summarizes the quality of the solutions obtained using the approximation and heuristic algorithms for small instances of the problem. The six columns presented in the table indicate the number of targets in the instances, the apriori ratio of the approximation algorithm, the average aposteriori ratios of the approximation and the heuristic algorithms, and the average ratios of the algmorithmic costs with the lower bound for the approximation and heuristic algorithms respectively. All the averages are taken over ($50$) instances with the same number of targets. }
    \centering
    \begin{tabular}{c|c|c|c|c|c}
    \toprule
    \multirow{2}{*}{\# Targets} & \multirow{2}{*}{Apriori Ratio} & \multicolumn{2}{|c|}{Avg. Aposteriori Ratio} &  \multicolumn{2}{|c|}{Avg. Ratio w..r.t. L.B.}\\
    \cmidrule{3-4}\cmidrule{5-6}
      & (Approx. Algo.) & Approx. Algo. & Heur. Algo.  & Approx. Algo. & Heur. Algo.\\
     \toprule
        6 & 3.75 & 1.12 & 1.13 & 1.39 & 1.40\\
        8 & 3.75 & 1.05 & 1.05 & 1.44 & 1.43\\
        10 & 3.75 & 1.07 & 1.05 & 1.47 & 1.44\\
        12 & 3.75 & 1.08 & 1.05 & 1.49 & 1.45\\
        14 & 3.75 & 1.11 & 1.06 & 1.55 & 1.44\\
        \bottomrule
    \end{tabular}
    \label{tab:avg-ratios-small-instances}
\end{table*}


\begin{table*}[]
    \caption{This table presents the results of numerical simulations performed over $50$ of the $250$ small instances. Here, the words aposteriori and lower bound are abbreviated as Apos. and L.B. respectively, in the interest of space. Note that the presented values are not the average values, unlike the results presented in other tables.}
    \centering
        \resizebox{\textwidth}{!}{\begin{tabular}{c|c|c|c|c|c|c|c|c|c|c|c|c|}
        \toprule
        \multirow{2}{*}{Instance \#} & \multirow{2}{*}{\# Targets} & \multicolumn{2}{|c|}{Optimal} & \multicolumn{4}{|c|}{Approximation Algorithm} & \multicolumn{4}{|c|}{Heuristic Algorithm} & \multirow{2}{*}{L.B} \\
        \cmidrule{3-4}\cmidrule{5-8}\cmidrule{9-12}
          &  & Value & Runtime & Value & Apos. Ratio & Runtime & Ratio w.r.t. L.B.&  Value & Apos. Ratio & Runtime & Ratio w.r.t. L.B.& \\
         \toprule
1  & 6 & 2019.58 & 0.58 & 2136.34 & 1.06 & 0.06 & 1.58 & 2140.62 & 1.06 & 0.05 & 1.59 & 1349.28 \\
2  & 6 & 2320.60 & 0.60 & 2631.67 & 1.13 & 0.06 & 1.39 & 2631.67 & 1.13 & 0.06 & 1.39 & 1891.15 \\
3  & 6 & 2340.35 & 0.59 & 2659.39 & 1.14 & 0.05 & 1.60 & 2994.48 & 1.28 & 0.06 & 1.8  & 1662.7  \\
4  & 6 & 1834.62 & 0.68 & 2081.49 & 1.13 & 0.06 & 1.37 & 2162.55 & 1.18 & 0.06 & 1.43 & 1515.15 \\
5  & 6 & 2375.09 & 0.69 & 2460.01 & 1.04 & 0.05 & 1.24 & 2507.83 & 1.06 & 0.06 & 1.26 & 1988.36 \\
6  & 6 & 2188.47 & 0.61 & 2728.47 & 1.25 & 0.06 & 1.51 & 2728.47 & 1.25 & 0.06 & 1.51 & 1807.81 \\
7  & 6 & 2411.82 & 0.55 & 2623.40 & 1.09 & 0.06 & 1.52 & 2669.99 & 1.11 & 0.06 & 1.54 & 1730.52 \\
8  & 6 & 1505.40 & 0.82 & 1613.71 & 1.07 & 0.05 & 1.22 & 1626.94 & 1.08 & 0.06 & 1.23 & 1323.43 \\
9  & 6 & 2015.54 & 0.78 & 2275.24 & 1.13 & 0.05 & 1.33 & 2309.27 & 1.15 & 0.06 & 1.35 & 1709.33 \\
10 & 6 & 1697.73 & 0.65 & 1819.00 & 1.07 & 0.05 & 1.41 & 1819.0  & 1.07 & 0.06 & 1.41 & 1294.15\\
         \midrule
11 & 8 & 2138.56 & 2.31 & 2289.06 & 1.07 & 0.06 & 1.60 & 2284.58 & 1.07 & 0.05 & 1.60 & 1427.70 \\
12 & 8 & 2934.60 & 2.27 & 3061.62 & 1.04 & 0.06 & 1.52 & 3061.62 & 1.04 & 0.06 & 1.52 & 2010.33 \\
13 & 8 & 2394.92 & 1.86 & 2626.79 & 1.10 & 0.05 & 1.32 & 2508.0  & 1.05 & 0.06 & 1.26 & 1986.35 \\
14 & 8 & 2768.99 & 2.08 & 2955.76 & 1.07 & 0.05 & 1.46 & 2955.76 & 1.07 & 0.06 & 1.46 & 2018.16 \\
15 & 8 & 2535.43 & 2.33 & 2584.59 & 1.02 & 0.06 & 1.48 & 2584.59 & 1.02 & 0.06 & 1.48 & 1746.54 \\
16 & 8 & 2453.88 & 1.88 & 2568.12 & 1.05 & 0.05 & 1.56 & 2568.12 & 1.05 & 0.06 & 1.56 & 1640.98 \\
17 & 8 & 1899.96 & 2.10 & 1967.28 & 1.04 & 0.05 & 1.38 & 1967.28 & 1.04 & 0.06 & 1.38 & 1425.23 \\
18 & 8 & 2299.77 & 1.81 & 2467.43 & 1.07 & 0.05 & 1.43 & 2467.43 & 1.07 & 0.06 & 1.43 & 1727.66 \\
19 & 8 & 1878.24 & 3.50 & 1912.46 & 1.02 & 0.06 & 1.57 & 1918.84 & 1.02 & 0.06 & 1.57 & 1221.31 \\
20 & 8 & 2236.81 & 2.56 & 2341.80 & 1.05 & 0.05 & 1.19 & 2567.15 & 1.15 & 0.06 & 1.30 & 1970.23\\
         \midrule
21 & 10 & 2543.27 & 11.43 & 2598.69 & 1.02 & 0.06 & 1.51 & 2598.69 & 1.02 & 0.06 & 1.51 & 1724.94 \\
22 & 10 & 2935.21 & 8.50  & 3357.22 & 1.14 & 0.06 & 1.61 & 3038.09 & 1.04 & 0.06 & 1.46 & 2084.44 \\
23 & 10 & 2535.19 & 7.61  & 2641.20 & 1.04 & 0.06 & 1.48 & 2641.2  & 1.04 & 0.06 & 1.48 & 1782.02 \\
24 & 10 & 3094.96 & 13.24 & 3236.95 & 1.05 & 0.06 & 1.45 & 3182.03 & 1.03 & 0.13 & 1.42 & 2233.05 \\
25 & 10 & 3073.28 & 8.23  & 3149.87 & 1.02 & 0.06 & 1.54 & 3149.87 & 1.02 & 0.05 & 1.54 & 2041.04 \\
26 & 10 & 3021.90 & 6.63  & 3149.63 & 1.04 & 0.05 & 1.38 & 3174.14 & 1.05 & 0.06 & 1.39 & 2284.17 \\
27 & 10 & 2782.15 & 8.12  & 3060.80 & 1.10 & 0.05 & 1.42 & 3060.8  & 1.1  & 0.06 & 1.42 & 2160.84 \\
28 & 10 & 2825.82 & 10.94 & 3147.48 & 1.11 & 0.06 & 1.49 & 3050.67 & 1.08 & 0.06 & 1.44 & 2116.82 \\
29 & 10 & 3057.25 & 8.05  & 3226.38 & 1.06 & 0.06 & 1.44 & 3226.38 & 1.06 & 0.06 & 1.44 & 2236.65 \\
30 & 10 & 2484.17 & 8.41  & 2497.13 & 1.01 & 0.06 & 1.59 & 2497.13 & 1.01 & 0.06 & 1.59 & 1571.97\\
         \midrule
31 & 12 & 3358.94 & 63.02 & 3404.75 & 1.01 & 0.06 & 1.59 & 3404.75 & 1.01 & 0.06 & 1.59 & 2143.79 \\
32 & 12 & 3209.36 & 25.09 & 3492.69 & 1.09 & 0.06 & 1.44 & 3422.02 & 1.07 & 0.06 & 1.41 & 2430.32 \\
33 & 12 & 3019.62 & 40.89 & 3168.65 & 1.05 & 0.06 & 1.44 & 3168.65 & 1.05 & 0.06 & 1.44 & 2198.50 \\
34 & 12 & 3048.37 & 82.02 & 3387.42 & 1.11 & 0.06 & 1.55 & 3197.42 & 1.05 & 0.06 & 1.46 & 2190.26 \\
35 & 12 & 2632.85 & 35.60 & 2850.74 & 1.08 & 0.06 & 1.45 & 2867.6  & 1.09 & 0.06 & 1.46 & 1962.80 \\
36 & 12 & 3571.70 & 41.85 & 4078.24 & 1.14 & 0.06 & 1.60 & 3679.15 & 1.03 & 0.06 & 1.45 & 2541.87 \\
37 & 12 & 3230.83 & 28.74 & 3671.91 & 1.14 & 0.06 & 1.49 & 3640.51 & 1.13 & 0.06 & 1.48 & 2460.07 \\
38 & 12 & 3109.34 & 52.93 & 3370.09 & 1.08 & 0.06 & 1.43 & 3363.59 & 1.08 & 0.07 & 1.42 & 2361.65 \\
39 & 12 & 2945.09 & 44.75 & 2993.71 & 1.02 & 0.06 & 1.48 & 3003.07 & 1.02 & 0.06 & 1.48 & 2027.05 \\
40 & 12 & 2943.53 & 35.41 & 3067.26 & 1.04 & 0.06 & 1.46 & 3096.23 & 1.05 & 0.06 & 1.48 & 2094.17\\
         \midrule
41 & 14 & 3162.88 & 227.96 & 3238.05 & 1.02 & 0.06 & 1.51 & 3238.05 & 1.02 & 0.06 & 1.51 & 2142.37 \\
42 & 14 & 3261.45 & 410.44 & 4058.65 & 1.24 & 0.06 & 1.82 & 3419.32 & 1.05 & 0.06 & 1.54 & 2225.74 \\
43 & 14 & 3550.04 & 152.61 & 4112.20 & 1.16 & 0.06 & 1.45 & 3742.91 & 1.05 & 0.06 & 1.32 & 2828.44 \\
44 & 14 & 3210.55 & 171.60 & 3522.19 & 1.10 & 0.06 & 1.45 & 3402.58 & 1.06 & 0.06 & 1.4  & 2428.94 \\
45 & 14 & 3136.03 & 120.96 & 3774.31 & 1.20 & 0.06 & 1.60 & 3245.69 & 1.03 & 0.06 & 1.38 & 2356.58 \\
46 & 14 & 3505.23 & 196.39 & 3748.90 & 1.07 & 0.06 & 1.48 & 3748.9  & 1.07 & 0.06 & 1.48 & 2525.91 \\
47 & 14 & 3037.45 & 210.29 & 3261.26 & 1.07 & 0.06 & 1.40 & 3229.29 & 1.06 & 0.06 & 1.38 & 2331.63 \\
48 & 14 & 2951.75 & 204.56 & 3409.17 & 1.15 & 0.06 & 1.64 & 3108.83 & 1.05 & 0.07 & 1.49 & 2082.47 \\
49 & 14 & 3400.72 & 190.60 & 3916.57 & 1.15 & 0.06 & 1.71 & 3687.0  & 1.08 & 0.06 & 1.61 & 2293.76 \\
50 & 14 & 3551.24 & 104.07 & 4044.55 & 1.14 & 0.06 & 1.47 & 3812.4  & 1.07 & 0.06 & 1.38 & 2760.70\\
         \bottomrule
    \end{tabular}}
    \label{tab:sample-values-small-instances}
\end{table*}

\subsection{Large Instances}
The number of targets in these instances range from $20$ to $100$. Due to the size of these instances, optimal solutions were not computable within the $3$-hour cut-off time. For every large instance, we compute the approximate and heuristic solutions, and determine the ratios of the costs of these solutions to the lower bounds. The average ratios and the average computation times for both the algorithms are presented in Table \ref{tab:large-instances}. 

\begin{table*}[]
\caption{This table summarizes the results of the simulations performed over the $250$ large instances of the problem. The five columns of the table indicate the number of targets in the instances, the average ratio of the cost of the algorithmic solution to the lower bound and the average computation time (runtime) for the approximation algorithm, followed by the corresponding values of the heuristic algorithm. The averages are taken over ($50$) instances with identical number of targets.}
\centering
    \begin{tabular}{|c|c|c|c|c|}
        \toprule
        \multirow{2}{*}{\# Targets} & \multicolumn{2}{|c|}{Approximation Algorithm} & \multicolumn{2}{|c|}{Heuristic Algorithm}\\
        \cmidrule{2-3}\cmidrule{4-5}
        & Avg. Ratio w.r.t. L.B. & Avg. Runtime (s) & Avg. Ratio w.r.t. L.B. & Avg. Runtime (s)\\
        \toprule
        20  & 1.54 & 0.06 & 1.48 & 0.06 \\
        30  & 1.57 & 0.07 & 1.48 & 0.08 \\
        40  & 1.59 & 0.08 & 1.49 & 0.15 \\
        50  & 1.59 & 0.09 & 1.49 & 0.21 \\
        100 & 1.61 & 0.27 & 1.50 & 0.89\\
        \bottomrule
    \end{tabular}
\label{tab:large-instances}
\end{table*}

The results suggest that the average ratios for both the algorithms increase with the number of targets in the instance. Nonetheless, similar to the trend observed in small instances, the rate of increase is smaller for the heuristic algorithm compared to its counterpart. While the ratios range from 1.54 to 1.61 for the approximation algorithm, they range from 1.48 to 1.50 for the heuristic algorithm. It is to be noted that these ratios are not significantly higher than those observed in the case of small instances. Therefore, it is likely that the quality of the solutions provided by these algorithms is as good as that observed in small instances. However, this claim cannot be justified without improved lower bounds in future.

\begin{figure}
    \centering
    \includegraphics[width=0.5\textwidth]{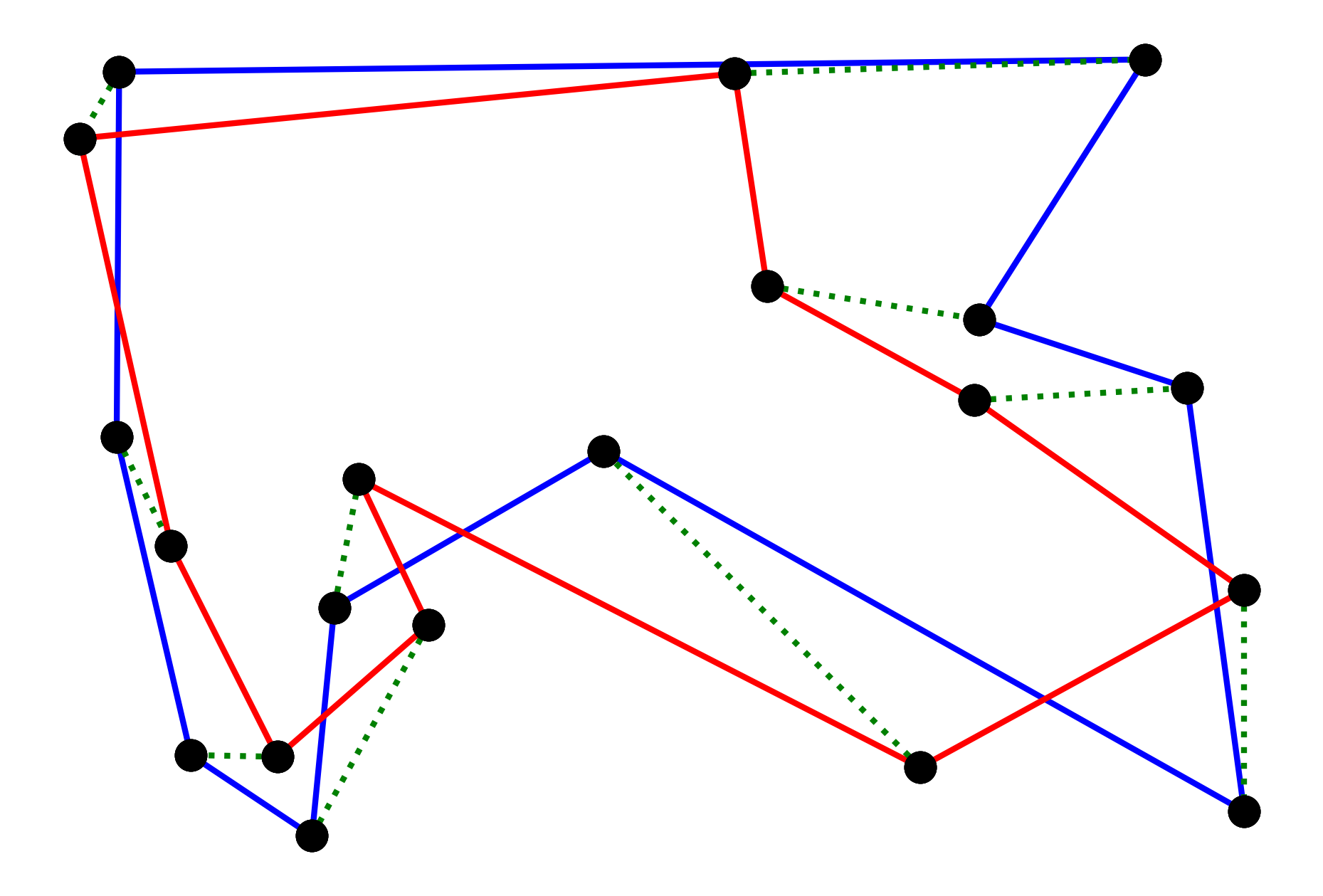}
    \caption{A feasible solution for the problem computed using the heuristic algorithm for a $20-$target instance. The black dots represent the target locations, the blue and red lines correspond to the $1^{st}$ and $2^{nd}$ vehicle tours, and the green dotted lines depict the communication links between the vehicles.}
    \label{fig:20ha}
\end{figure}

\begin{figure}
    \centering
    \includegraphics[width=0.5\textwidth]{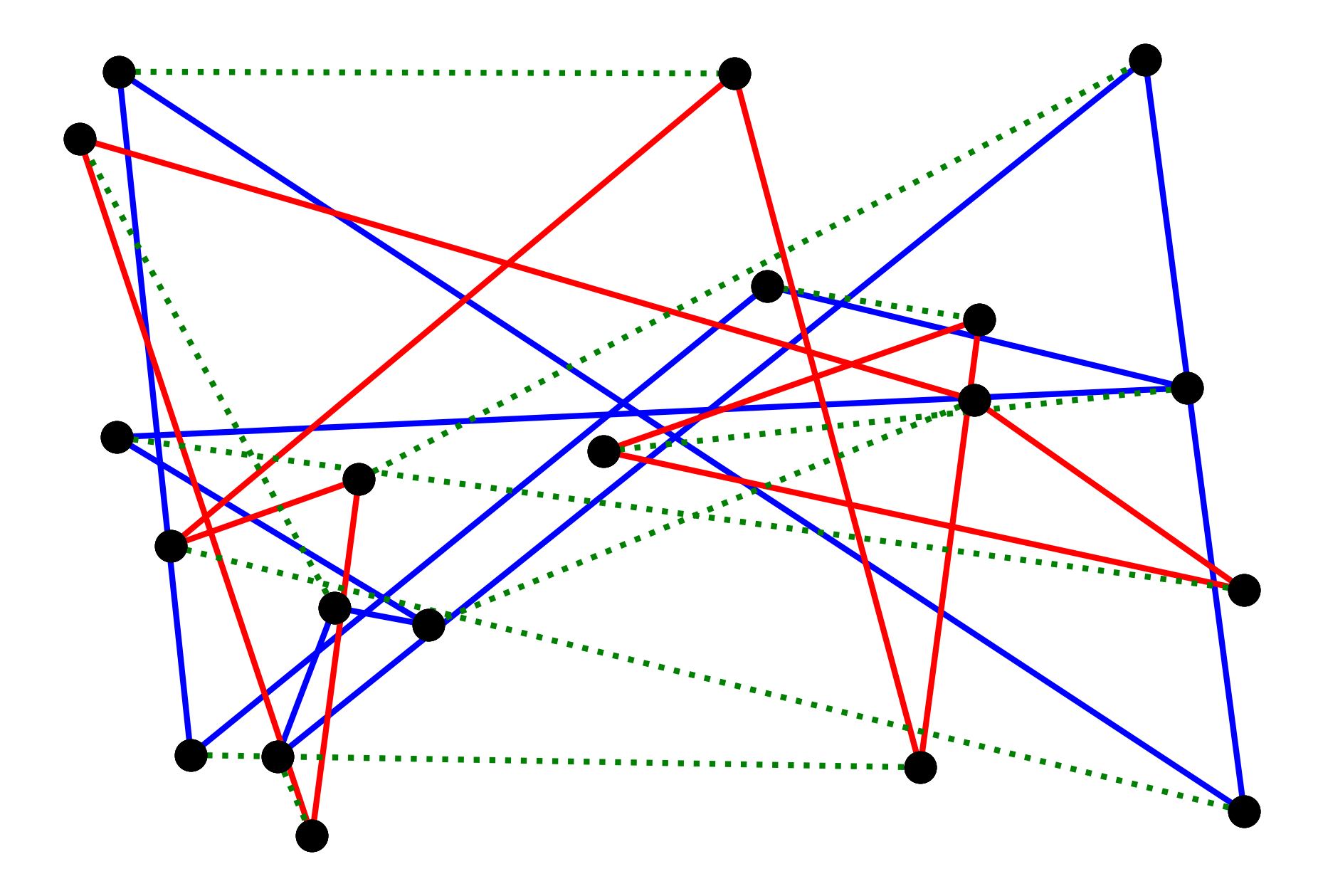}
    \caption{A feasible solution for the problem computed using the approximation algorithm for a $20-$target instance. The black dots represent the target locations, the blue and red lines correspond to the $1^{st}$ and $2^{nd}$ vehicle tours, and the green dotted lines depict the communication links between the vehicles.}
    \label{fig:20aa}
\end{figure}

In comparison, the quality of the solutions provided by the heuristic algorithm was observed to be slightly better than those provided by the approximation algorithm. This can be attributed to the difference in the type of the single UV tours provided by these algorithms. Figures \ref{fig:20ha} and \ref{fig:20aa} depict feasible solutions for an illustrative $20-$target instance computed using the heuristic and the approximation algorithms respectively. As can be seen from the figure, the vehicle tours provided by the latter contain criss-cross paths as opposed to those provided by the former. The presence of criss-crossing is an undesirable feature in TSP tours and contributes to higher costs.

\begin{figure}
    \centering
    \includegraphics[width=0.5\textwidth]{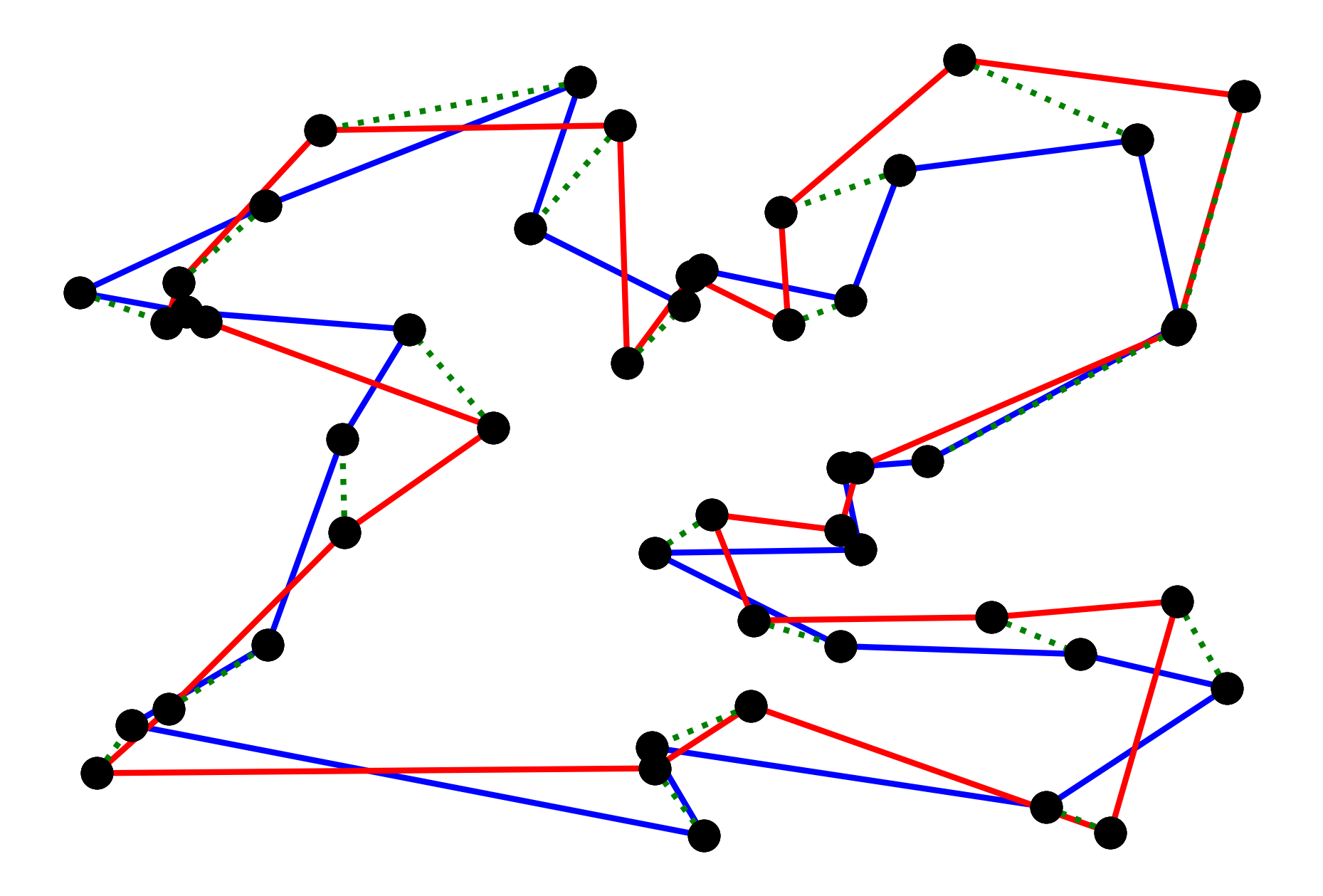}
    \caption{A feasible solution for the problem computed using the heuristic algorithm for a $50-$target instance. The black dots represent the target locations, the blue and red lines correspond to the $1^{st}$ and $2^{nd}$ vehicle tours, and the green dotted lines depict the communication links between the vehicles.}
    \label{fig:50ha}
\end{figure}

\begin{figure}
    \centering
    \includegraphics[width=0.5\textwidth]{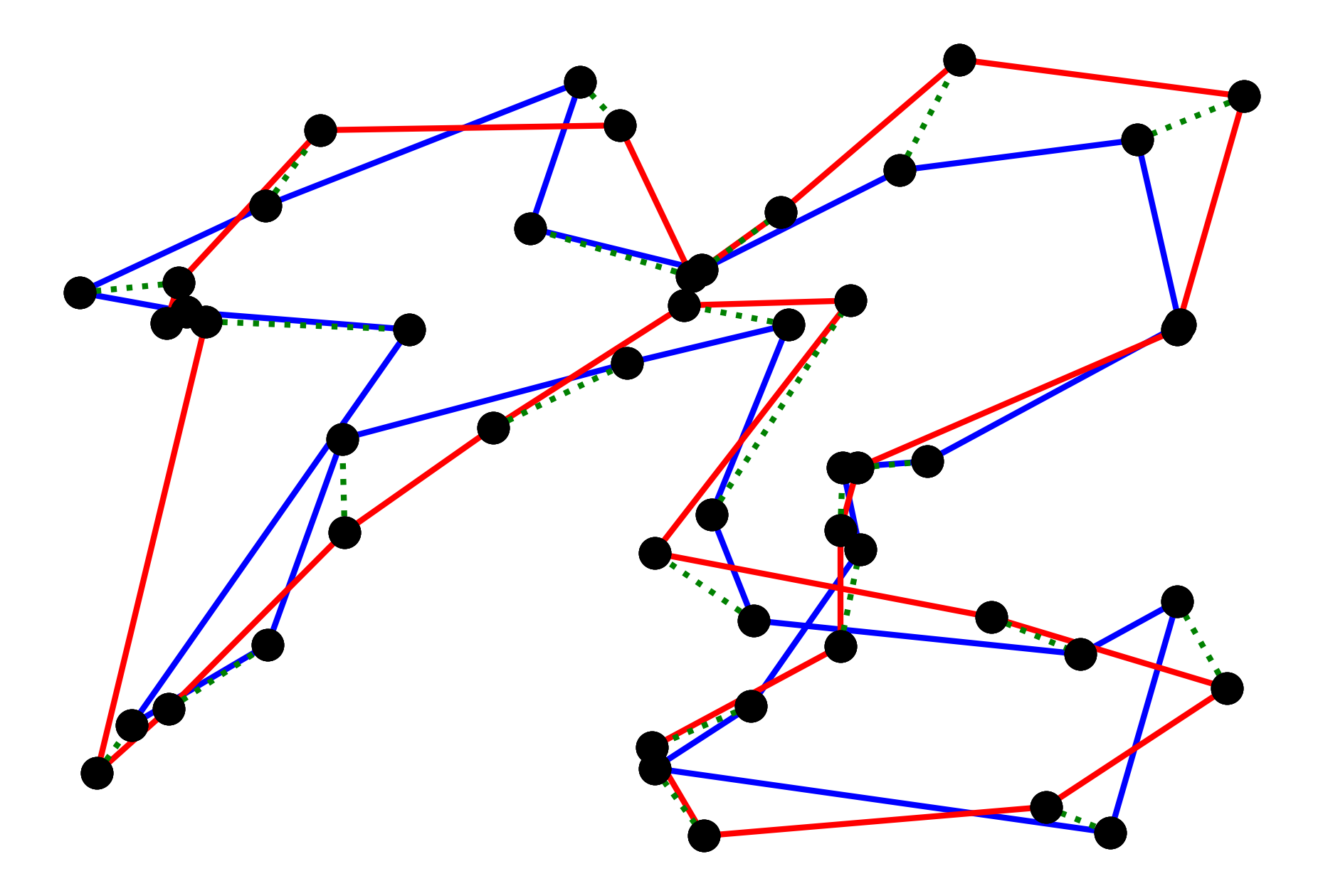}
    \caption{A feasible solution for the problem computed using the approximation algorithm for a $50-$target instance. The black dots represent the target locations, the blue and red lines correspond to the $1^{st}$ and $2^{nd}$ vehicle tours, and the green dotted lines depict the communication links between the vehicles.}
    \label{fig:50aa}
\end{figure}

However, the computation time of the approximation algorithm was observed to be slightly better than that of the heuristic algorithm for large instances. Therefore, there is a trade-off involved in selecting the desired algorithm. Nevertheless, the computation times for both the algorithms are within a fraction of a second for instances with upto $100$ targets. Therefore, the algorithms provide good-quality solutions quickly, and are useful in practice, especially when optimal solutions are not computable within the required time. Figures \ref{fig:50ha} and \ref{fig:50aa} show feasible solutions computed in 0.20 and 0.08 seconds for a $50-$target instance using the heuristic and approximation algorithms respectively.

\section{Summary}

We proposed a coordinated routing problem in which two unmanned vehicles, a leader and a wingmate, must provide a cooperative coverage of the environment while staying closed to each other. The objective of the problem is to find travel routes and communication links for the vehicle such that the sum of the travel and communication costs of the mission is minimized. Finding optimal solutions to large instances of the problem is computationally difficult, and therefore, we proposed approximation and heuristic algorithms to solve the problem swiftly. In practice, the approximation algorithm performed better than its predicted worst-case bound. Nonetheless, the costs of the solutions provided by the heuristic algorithm were observed to be marginally better that its counterpart. Besides, both the algorithms provided feasible solutions to the problem within a fraction of a second, for instances with upto $100$ targets on a MacBook Pro with 16 GB RAM, Quad-Core Intel Core i7 processor @ 2.8 GHz.

For small instances of the problem, where optimal solutions were available, the cost of the solutions provided by the heuristic algorithm were approximately $7 \%$ away from the optimal values, on an average. To evaluate the quality of the solutions for large instances where optimal solutions are unavailable, we developed lower bounds to the problem. Simulations suggest that the gap between the solution costs and the lower bounds were similar for large and small instances. However, future work calls for the development of improved lower bounds to the problem, which help in evaluating the quality of the algorithms developed in this paper or the ones that might be developed in future, for large instances.

\bibliographystyle{IEEEtran}
\bibliography{references.bib}
\end{document}